\documentclass[a4paper,twoside]{article}

\usepackage{epsfig}
\usepackage{subfigure}
\usepackage{calc}
\usepackage{amssymb}
\usepackage{amstext}
\usepackage{amsmath}
\usepackage{amsthm}
\usepackage{multicol}
\usepackage{multirow}
\usepackage{pslatex}
\usepackage{apalike}
\usepackage{SCITEPRESS}
\usepackage[small]{caption}
\usepackage{comment}
\usepackage[usenames,dvipsnames]{color}
\usepackage{multicol}
\usepackage{pdfpages}

\usepackage{fancyhdr}
\definecolor{orange}{rgb}{1,0.5,0}
\definecolor{darkgreen}{rgb}{0.1,0.5,0}
\definecolor{Mycolor}{rgb}{0.2,0.3,1}

\newcommand{\Esconde}[1]{}

\newcommand{\FFDnR}{FFireDt}
\newcommand{\ID}{image\ descriptor}

\newcommand{\FlickrFire}{Flickr-Fire}

\newtheorem{definition}{Definition}

\begin{document}

\title{Techniques for effective and efficient fire \\ detection from social media images}

\author{\authorname{Marcos Bedo, Gustavo Blanco, Willian Oliveira, Mirela Cazzolato, Alceu Costa, Jose Rodrigues, Agma Traina and Caetano Traina}
\affiliation{Institute of Mathematics and Computer Science, University of S\~{a}o Paulo \\ Av. Trabalhador S\~{a}o Carlense, 400, S\~{a}o Carlos, SP, Brazil}
\email{\{bedo, willian, alceufc, junio, agma, caetano\}@icmc.usp.br, \{blanco, mirelac\}@usp.br}
}


\abstract{
Social media could provide valuable information to support decision making in crisis management, such as in accidents, explosions and fires.
However, much of the data from social media are images, which are uploaded in a rate that makes it impossible for human beings to analyze them.
Despite the many works on image analysis, there are no fire detection studies on social media.
To fill this gap, we propose the use and evaluation of a broad set of content-based image retrieval and classification techniques for fire detection.
Our main contributions are:
(\textit{i}) the development of the Fast-Fire Detection method ($\FFDnR$), which combines feature extractor and evaluation functions to support instance-based learning;
(\textit{ii}) the construction of an annotated set of images with ground-truth depicting fire occurrences -- the \FlickrFire ~dataset; and
(\textit{iii}) the evaluation of 36 efficient image descriptors for fire detection.
Using real data from Flickr, our results showed that $\FFDnR$ was able to achieve a precision for fire detection comparable to that of human annotators.
Therefore, our work shall provide a solid basis for further developments on monitoring images from social media.
}

\onecolumn \maketitle \normalsize \vfill

\section{\uppercase{Introduction}}\label{sec:introduction}
\noindent
		Disasters in industrial plants, densely populated areas, or even crowded events may impact property, environment, and human life.
		For this reason, a fast response is essential to prevent or reduce injuries and financial losses, when crises situations strike.
		The management of such situations is a challenge that requires fast and effective decisions based on the best data available, because decisions based on incorrect or lack of information may cause more damage \cite{Russo2013}.
		One of the alternatives to improve information correctness and availability for decision making during crises is the use of software systems to support experts and rescue forces \cite{Kudyba2014}.

		Systems aimed at supporting salvage and rescue teams often rely on images to understand the crisis scenario and to design the actions that will reduce losses.
		Crowdsourcing and social media, as massive sources of images, possess a great potential to assist such systems. 
		Web sites such as Flickr, Twitter, and Facebook allow users to upload pictures from mobile devices, what generates a flow of images that carries valuable information. 
		Such information may reduce the time spent to make decisions and it can be used along with other information sources. 
		Automatic image analysis is important to understand the dimensions, type, and the objects and people involved in an incident.

		Despite the potential benefits, we observed that there is still a lack of studies concerning automatic content-based processing of crisis images \cite{Villela2014}.
		In the specific case of fire -- which is observed during explosions, car accidents, forest and building fire, to name a few, there is an absence of studies to identify the most adequate content-based retrieval techniques (image descriptors) able to identify and retrieve relevant images captured during crises. 
		In this work, we fill one of the gaps providing an architecture and an evaluation of the techniques for fire monitoring in images collected from social media.

This work reports on one of the steps of the project {\it Reliable and Smart Crowdsourcing Solution for Emergency and Crisis Management -- Rescuer}\footnote{http://www.rescuer-project.org/}.
		The project goal is to use crowd-sourcing data (image, video, and text captured with mobile devices) to assist in rescue missions.
		This paper describes the evaluation of techniques to detect fire in image data, one of the project targets.
		We use real images from Flickr\footnote{https://www.flickr.com/}, a well-known social media website from where we collected a large set of images that were manually annotated as having fire or not fire.
		We used this dataset as a ground-truth to evaluate image descriptors in the task of detecting fire.
		Our main contributions are the following:

\begin{enumerate}

\item \textbf{Curation of the \FlickrFire ~Dataset:} a vast human-annotated dataset of real images suitable as ground-truth for the development of content-based techniques for fire detection;

\item \textbf{Development of FFireDt:} we propose the Fast-Fire Detection and Retrieval ($\FFDnR$), a scalable and accurate architecture for automatic fire-detection, designed over image descriptors able to detect fire, which achieves a precision comparable to that of human annotation;

\item \textbf{Evaluation:} we soundly compare the precision and performance of several image descriptors for image classification and retrieval.
\end{enumerate}

		Our results provide a solid basis to choose the most adequate pair feature extractor and evaluation function in the task of fire detection, as well as a comprehensive discussion of how the many existing alternatives work in such task.

		The remaining of this paper is structured as follows: Section~\ref{sec:relatedwork} presents the related work; Section~\ref{sec:background} presents the main concepts regarding fire-detection in images; Section~\ref{sec:methods} presents the methodology. Section~\ref{sec:experiments} describes the experiments and discusses their results; finally, Section~\ref{sec:conclusion} presents the conclusions.

\section{\uppercase{Related Work}}\label{sec:relatedwork}
		Previous efforts on mining information from sets of images include detecting social events and tracking the corresponding related topics which can even include the identification of touristic attractions \cite{Tamura2012}.

		Distinctly, in this paper we are interested in the following problem: 
				\textit{Given a collection of photos, possibly obtained by a social media service, how can we efficiently detect fire?}
		Interesting approaches related to fire motion analysis on video are not applicable for static images \cite{Chunyu2010}, and most of these approaches where found not to work with satisfactory performance \cite{Celik2007,Ko2009,Liu2004}.

		Some of the previous works propose the construction of a particular color model focused on fire, based on Gaussian differences \cite{Celik2007} or in spectral characteristics to identify fire, smoke, heat or radiation. 
		The spectral color model has been used along with spatial correlation and a stochastic model to capture fire motion \cite{Liu2004}. However, such technique requires a set of images and is not suitable for individual images, as it is frequent with social media.

		Other studies employ a variation of the combination given by a color model transform plus a classifier.
		This combination is employed  in the work of Dimitropoulos \cite{Dimitropoulos2014}, which represents each frame according to the most prominent texture and shape features.
		It also combines such representation with spatio-temporal motion features to employ SVM to detect fire in videos.
		However, this approach is neither scalable nor suitable for fire detection on still images.

		On the other hand, the feature extraction methods available in the MPEG-7 Standard have been used for image representation in fast-response systems that deal with large amounts of data \cite{Doeller2008,Ojala2002,Tjondronegoro2002}. 
		However, to the best of our knowledge, there is no study employing those extractors for fire detection.
		Moreover, despite these multiple approaches, there is no conclusive work about which image descriptors are suitable to identify fire in images.

\section{\uppercase{Background}}\label{sec:background}

\subsection{Content-based model for retrieval and classification}

		The most usual approach to recover and classify images by content relies on representing them using feature extractor techniques \cite{Guyon2006}. 
		After extraction, the images can be retrieved comparing their feature vectors using an evaluation function, which is usually a metric or a divergence function.
		Comparison is a required step in image retrieval systems. Also, Instance-Based Learning (IBL) classifiers use the evaluation among the images of a given set to label them regarding their visual content \cite{Bedo2014,Aha1991}. 
		Such concepts can be formalized by the following definitions.

\begin{definition}
{\normalfont \textbf{Feature Extraction Method (FEM): }}
A feature extraction method is a non-bijective function that, given an image domain $\mathbb{I}$, is able to represent any image $i_q \in \mathbb{I}$ in a domain $\mathbb{F}$ as $f_q$.
		Each value $f_q$ is called a \textit{feature vector} (FV) and represents characteristics of an image $i_q$.
\end{definition}

		In this paper, we use FEMs to represent images in multidimensional domains. Therefore, the image feature vectors can be compared according to the next definition:

\begin{definition}\label{def:df}
{\normalfont \textbf{Evaluation Function (EF): }}
Given the feature vectors $f_i$, $f_j$ and $f_k \in \mathbb{F}$, an evaluation function $\delta ~: ~\mathbb{F} \times \mathbb{F} \rightarrow \mathbb{R}$ is able to compare any two elements from $\mathbb{F}$.
The EF is said to be a metric distance function if it complies with the following properties:
\begin{itemize}
	\item Symmetry: $\delta(f_i, f_j) = \delta(f_j, f_i)$.
	\item Non-negativity: $0 < \delta(f_i, f_j) < \infty$.
	\item Triangular inequality: $\delta(f_i, f_j) \leq \delta(f_i, f_k) + \delta(f_k, f_j)$.
\end{itemize}
\end{definition}

		The FEM defines the element distribution in the multidimensional space.
		On the other hand, the evaluation function defines the behavior of the searching functionalities.
		Therefore, the combination of FEM and EF is the main parameter to improve or decrease accuracy and quality for both classification and retrieval.
		Formally, this association can be defined as:

\begin{definition}
{\normalfont \textbf{Image Descriptor (ID): }}
An image descriptor is a pair $<\epsilon, ~\delta>$, where $\epsilon$ is a (composition of) FEM and $\delta$ is a (weighted) EF.
\end{definition}
		By employing a suitable $\ID$, it is possible to inspect the neighborhood of a given element considering previous labeled cases.
		This course of action is the principle of the Instance-Based Learning algorithms, which rely on previously labeled data to classify new elements according to their nearest neighbors. 
		The sense of what ``nearest'' means is provided by the EF.
		Formally, this operation must respect the definition bellow:

\begin{definition}
{\normalfont \textbf{k Nearest-Neighbors - kNN: }}
		Given an image $i_q$ represented as $f_q \in \mathbb{F}$, an image descriptor ID $= <\epsilon, ~\delta>$, a number of neighbors $k \in \mathbb{N}$ and a set $F$ of images, the $k$-Nearest Neighbors set is the subset of $F \subset \mathbb{F}$ such that $kNN = \{f_n \in \mathbb{F} ~| ~ \forall ~f_i \in F; ~\delta(f_n, ~f_q) < \delta(f_i, ~f_q) \}$.
\end{definition}

		Once the $kNN$ performance relies on the capability of the $\ID$ to define the image representations and the search space, it becomes the critical point to be defined in an image retrieval system.
		Following we review, experiment, and report on multiple possibilities of image descriptors for fire detection.

\subsection{MPEG-7 Feature Extraction Methods}
\label{subsec:mpeg7fems}

		The MPEG-7 standard was proposed by the ISO/IEC JTC1 \cite{IEEE2002}. 
		It defines expected representations for images regarding color, texture and shape. 
		The set of proposed feature extraction methods were designed to process the original image as fast as possible, without taking into account specific image domains.
		The original proposal of MPEG-7 is composed of two parts: high and low-level values, both intended to represent the image.
		The low-level value is the representation of the original data by a FEM. 
		On the other hand, the high-level feature requires examination by an expert.

		The goal of MPEG-7 is to standardize the representation of \textit{streamed} or stored images.
		The low-level FEMs are widely employed to compare and to filter data, based purely on content.
		These FEMs are meaningful in the context of various applications according to several studies \cite{Doeller2008,Tjondronegoro2002}.
		They are also supposed to define objects by including color patches, shapes or textures. 
		The MPEG-7 standard defines the following set of low-level extractors \cite{Sato2010}:

\begin{itemize}
	\item \textbf{Color:} Color Layout, Color Structure, Scalable Color and Color Temperature, Dominant Color, Color Correlogram, Group-of-Frames;
	\item \textbf{Texture:} Edge Histogram, Texture Browsing, Homogeneous Texture;
	\item \textbf{Shape:} Contour Shape, Shape Spectrum, Region Shape;
\end{itemize}

		We highlight that shape FEMs' usually depends of previous object definitions.
		As the goal of this study relies on defining the best setting for automatic classification and retrieval for fire detection, user interaction on the extraction process is not suitable for our proposal.
		Thus, we focus only on the color and texture extractors.
		In this study we employ the following MPEG-7 extractors: 
				Color Layout, Color Structure, Scalable Color, Color Temperature, 
				Edge Histogram, and Texture Browsing. 
		They are explained in the next Sections.

\subsubsection{Color Layout}

\begin{figure}[ht]
\centering
\includegraphics[scale=0.71]{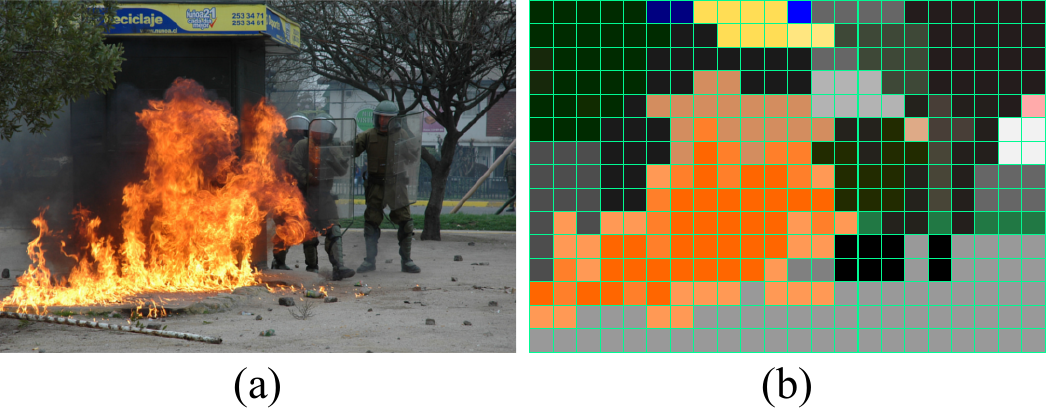}
\caption{(a) Original image (b) Regions considered by Color Layout.}
\label{fig:colorlayout}
\end{figure}

		The MPEG-7 Color Layout (CL) \cite{Kasutani2001} describes the image color distribution considering spatial location. 
		It splits the image in squared sub-regions (the number of sub regions is a parameter) and label each square with the average color of the region.
		Figure \ref{fig:colorlayout}(b) depicts the regions for Figure \ref{fig:colorlayout}(a) according to the Color Layout extractor.
		Next, the average colors are transformed to the YCbCr space and a Discrete Cosine Transformation is applied over each band of the YCbCr region values.
		The low-frequency coefficients are extracted through a zig-zag image reading.
		In order to reduce dimensionality, only the most prominent frequencies are employed in the feature vector.

\subsubsection{Scalable Color}

		The MPEG-7 Scalable Color (SC) \cite{Manjunath2001} aims at capturing the prominent color distribution. It is based on four stages.
		The first stage converts all pixels from the RGB color-space to the HSV space and a normalized color histogram is constructed.
		The color histogram is quantized using 256 levels of the HSV space.
		Finally, a Haar wavelet transformation is applied over the resulting histogram \cite{Ojala2002}.

\subsubsection{Color Structure}

\begin{figure}[ht]
\centering
\includegraphics[scale=0.9]{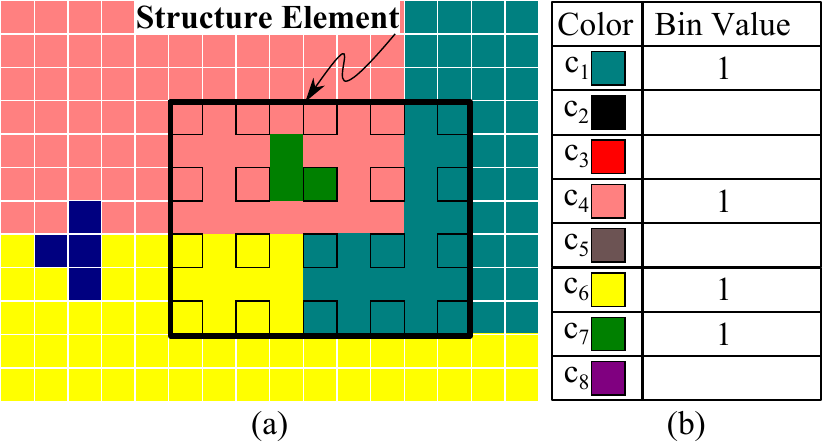}
\caption{(a) Color structure with a defined window (b) Local histogram.}
\label{fig:colorstructure}
\end{figure}

		The MPEG-7 Color Structure (CS) expresses both spatial and color distribution \cite{Sikora2001}.
		This paper splits the original image in a set of color structures with fixed-size windows.
		Each fixed-size window selects equally spaced pixels to represent the local color structure, as depicted in Figure \ref{fig:colorstructure}(a).

		The window size and the number of local structures are parameters of CS \cite{Manjunath2001}.
		For each color structure, a quantization based on the HMMD - a color-space derived from HSV that represents color differences - is executed.
		Then a local ``histogram'' based on HMMD is built.
		It stores the presence or absence of the quantized color instead of its distribution along with the window (Figure \ref{fig:colorstructure}(b)).
		The resulting feature vector is the accumulated distribution of the local histograms according to the previous quantization.

\subsubsection{Edge Histogram}

\begin{figure}[ht]
\centering
\includegraphics[scale=0.9]{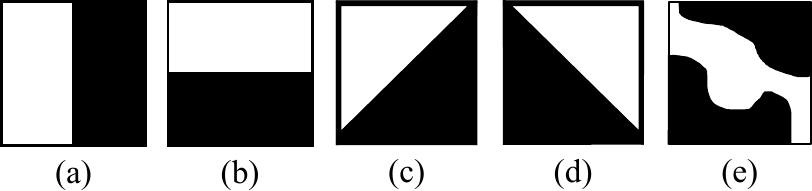}
\caption{Edge types: (a) Vertical (b) Horizontal (c) 45 degree (d) 135 degree (e) non-directional.}
\label{fig:edgeblock}
\end{figure}

		The MPEG-7 Edge Histogram (EH) aims at capturing local and global edges.
		It defines five types of edges (Figure \ref{fig:edgeblock}) regarding $N \times N$ blocks, where $N$ is a extractor parameter.
		Each block is constructed by partitioning the original image into squared regions.

\begin{figure}[ht]
\centering
\includegraphics[scale=0.9]{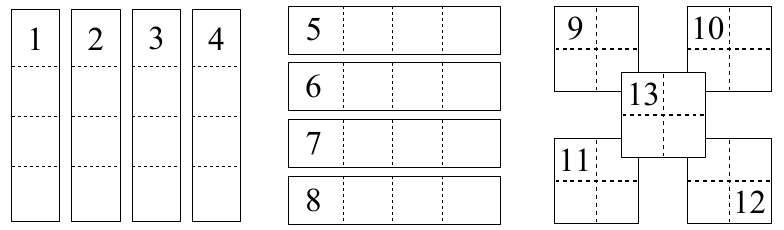}
\caption{13 regions corresponding to semi-local edge histograms.}
\label{fig:edgehistogram}
\end{figure}

		After applying the masks shown in Figure \ref{fig:edgeblock} to an image, it is possible to compute the local edge histograms.
		At this stage, the entire histogram is composed of $5 \times N$ bins, but it is biased by local edges.
		To circumvent this problem, a variation \cite{Park2000} was proposed to capture also semi-local edges.
		Figure \ref{fig:edgehistogram} illustrates how the 13 semi-local edges are calculated.
		The horizontal semi-local edges are evaluated first, then the vertical ones and finally the five combinations of the super block edges.

		The resulting feature vector is composed of $N$ plus thirteen edge-histograms, which represents the local and the semi-local distribution, respectively.

\subsubsection{Color Temperature}

\begin{figure}[ht]
\centering
\includegraphics[scale=0.6]{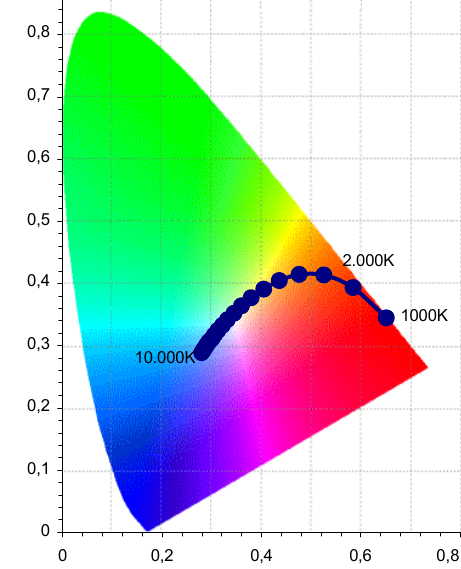}
\caption{CIE color system and black body locus indicated by the in-out points.}
\label{fig:colortemperature}
\end{figure}

		The main hypothesis supporting the MPEG-7 Color Temperature (CT) is that there is a correlation between the ``feeling of image temperature'' and illumination properties.
		Formally, the proposal considers a theoretical object called \textit{black body}, whereupon its color depends on the temperature \cite{Wnukowicz2003}.
		Figure \ref{fig:colortemperature} depicts the locus of the theoretical black body, according to Planck formula changing from 2000 Kelvin (red) to 25000 Kelvin (blue).

		The feature vector represent the linearized pixels in the XYZ space.
		This is performed by interactively discarding every pixel with luminance Y above the given threshold -- a FEM's parameter.
		Thereafter, the average color coordinate in XYZ is converted to UCS.
		Finally, the two closest isotemperature lines is calculated from the given color diagrams \cite{Wnukowicz2003}.
		The formula for the resulting color temperature depends on the average point, the closest isolines and the distances among them.

\subsubsection{Texture Browsing}

\begin{figure}[ht]
\centering
\includegraphics[scale=0.77]{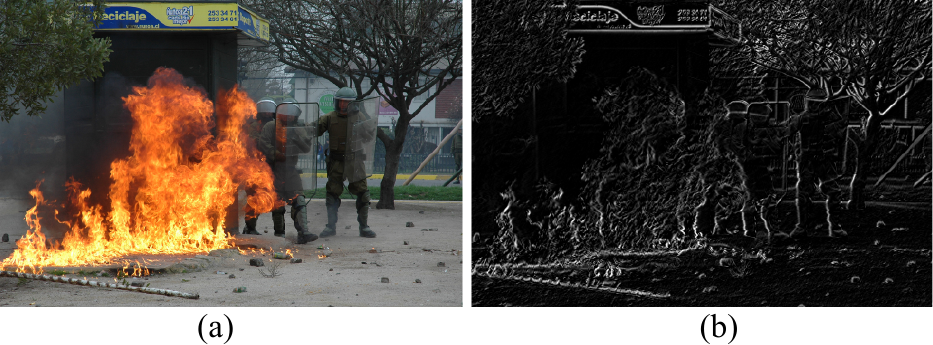}
\caption{(a) Original Image (b) Gabor filter with a kernel setting.}
\label{fig:gaborfilter}
\end{figure}

		The MPEG-7 Texture Browsing extractor (TB) is obtained from Gabor filters applied to the image \cite{Lee2005}.
		This FEM parameters' are the same used in Gabor filtering.
		Figure \ref{fig:gaborfilter} (b) ilustrates the result of using the Gabor filter to process image \ref{fig:gaborfilter} (a) following a particular setting.
		The Texture Browsing feature vector is composed of 12 positions: 2 to represent regularity, 6 for directionality and 4 for coarseness.

		The regularity features represent the degree of regularity of the texture structure as a more/less regular pattern, in such a way that the more regular a texture, the more robust the representation of the other features is.
		The directionality defines the most dominant texture orientation.
		This feature is obtained providing an orientation variation for the Gabor filters.
		Finally, the coarseness represents the two dominant scales of the texture.

\subsection{Evaluation Functions}

	An Evaluation Function expresses the proximity between two feature vectors.
	We are interested in feature extractor that generates the same amount of features for each image, thus in this paper we account only for evaluation functions for multidimensional spaces.
	Particularly, we employed distance functions (metrics) and divergences as evaluation functions.
	Suppose two feature vectors $X = \{x_1, ~x_2, ~\ldots, ~x_n\}$ and $Y = \{y_1, ~y_2, ~\ldots, ~y_n\}$ of dimensionality $n$.
	Table \ref{tab:df_form} shows the EFs implemented, according to their evaluation formulas.

\begin{table}[ht]
\caption{Evaluation functions: their classification as metric distance functions and respective formulas.}
\begin{center}
\begin{tabular}{|l|c|l|}
\hline
\hline
\textbf{Name}  & \textbf{Metric} & \textbf{Formula} \\ 
\hline
\hline
\small City-Block   & \small Yes    & $\sum^n_{i = 1}|x_i - y_i|$       \\ \hline
\small Euclidean    & \small Yes    & $\sqrt{\sum^n_{i = 1} (x_i - y_i)^2)} $      \\ \hline
\small Chebyshev    & \small Yes   &  $\lim_{p \to \infty} (\sum^n_{i = 1} |x_i - y_i|^p)^{\frac{1}{p}}$       \\ \hline
\small Canberra     & \small Yes  &  $\sum^n_{i = 1} \frac{| ~x_i - y_i ~|}{|x_i| + |y_i|} $      \\ \hline
\begin{tabular}[c]{@{}l@{}}\small Kullback\\\small  Leibler\\\small  Divergence\end{tabular} & \small No  & $\sum^n_{i = 1} x_i \ln(\frac{x_i}{y_i})$   \\ \hline
\begin{tabular}[c]{@{}l@{}}\small  Jeffrey\\\small  Divergence\end{tabular}            &\small  No  & $\sum^n_{i = 1} (x_i - y_i) \ln(\frac{x_i}{y_i})$  \\ \hline
\end{tabular}
\end{center}
\label{tab:df_form}
\end{table}

		The most widely employed metric distance functions are those related to the Minkowski family: the Manhattan, Euclidean and Chebyshev  \cite{Zezula2006}.
		A variation of the Manhattan distance is the Canberra distance that results in distances in the range $[0,1]$.
		These four EFs satisfy the properties of Definition \ref{def:df}.
		Therefore, they are metric distance functions.

		However, there are non-metric distance functions that are useful for image classification and retrieval.
		The Kullback-Leibler Divergence, for instance,  does not follow the triangular inequality neither the symmetry properties.
		A symmetric variation of Kullback-Leibler distance is the Jeffrey Divergence, yet it still is not a metric due to the lack of the triangular inequality compliance.

\subsection{Instance-Based Learning - IBL}
\label{subsec:ibl}
\noindent
		The main hypothesis for IBL classification is that the unlabeled feature vectors (FV) pertain to the same class of its $k$ Nearest-Neighbors, according to a predefined rule.
		Such classifier relies on three resources:
		\begin{enumerate}
			 \item An evaluation function, which evaluates the proximity between two FVs;
			 \item A classification function, which receives the nearest FVs to classify the unlabeled one -- commonly considering the majority of retrieved FVs;
			 \item A concept description updater, which maintains the record of previous classifications.
		\end{enumerate}

		Variation of these parts defines different IBL versions.
		For instance, the IB1 -- probably the most widely adopted IBL algorithm -- adopts the majority of the retrieved elements as the classification rule and keeps no record of previous classifications.

		The $kNN$ process is able to solve all steps required by IB1.
		Moreover, it can be seamlessly integrated to the concept of similarity queries by employing extended-SQL expressions \cite{Bedo2014}.
		This database-driven approach to solve IB1 may reduce the time to obtain the final classification by orders of magnitude, besides the obvious gains obtained by structuring queried data following the entity-relationship model.

\section{\uppercase{Proposed Method}}\label{sec:methods}

\subsection{Dataset \FlickrFire}\label{sec:dataset}
\begin{figure}[ht]
\centering
\includegraphics[scale=0.72]{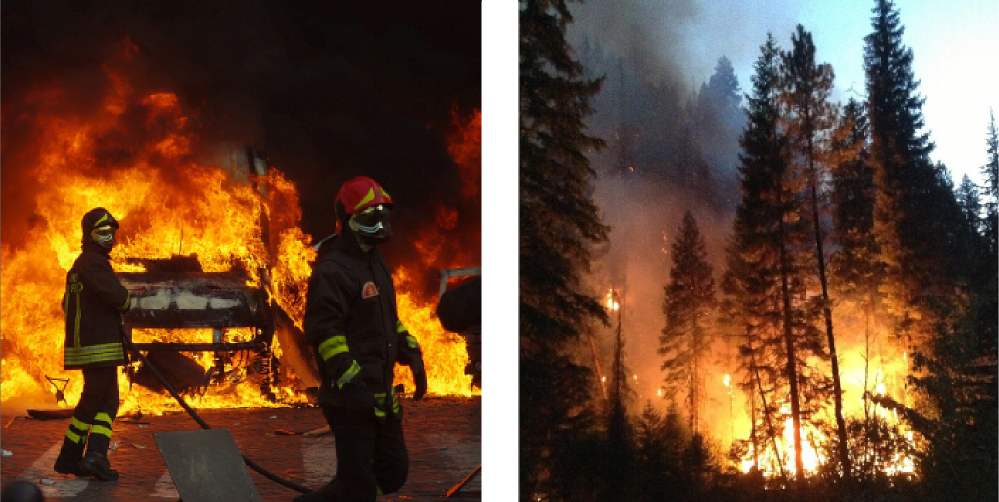}
\caption{Sample images labeled as 'fire' from dataset \texttt{\FlickrFire}.}
\label{fig:sample_fire}
\end{figure}

\begin{figure}[ht]
\centering
\includegraphics[scale=0.72]{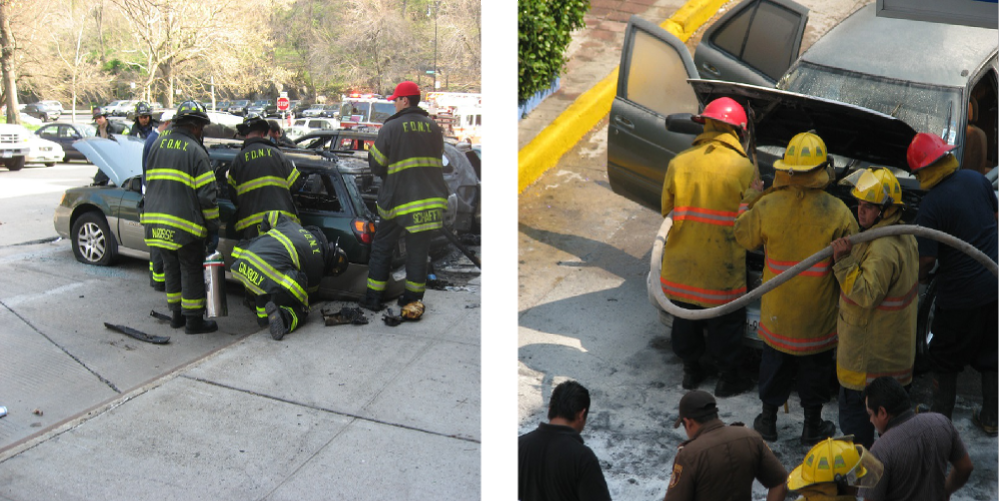}
\caption{Sample images labeled as 'not-fire' from dataset \texttt{\FlickrFire}.}
\label{fig:sample_not_fire}
\end{figure}

\noindent
		We used the Flickr API\footnote{The Flickr API is available at: \texttt{www.flickr.com/services/api/}} to download 5,962 images (no duplicates) under the Creative Commons license. 
		The images were retrieved using textual queries such as: ``fire car accident'', ``criminal fire'', and ``house burning''.
		Figures \ref{fig:sample_fire} and \ref{fig:sample_not_fire} illustrate samples of the obtained images, which we named \texttt{\FlickrFire}.
		Even with queries related to fire, some of the images did not contain visual traces of fire, so each image was manually annotated to define a coherent ground-truth dataset.

		To perform the annotation, we asked 7 subjects, all of them aging between 20 and 30 years, familiar with the issue, and non-color-blinded.
		To each subject it was given a subset with 1,589 images that he/she should annotate as containing or not traces of fire. 
		For images in which the annotations disagreed, we asked a third subject to provide an annotation. 
		The average disagreement was 7.2\%.

		In order to balance the class distribution of the dataset, we randomly removed images to have 1,000 images containing fire and 1,000 images without fire. 
		We made the dataset available online\footnote{The Flickr-Fire dataset at: \texttt{www.gbdi.icmc.usp.br}} aiming at the reproducibility of our experiments.

\subsection{The Architecture of Fast-Fire Detection}
\noindent

\begin{table}[!htb]
\caption{Feature Extractor Method acronyms used in the experiments.}
\begin{center}
\begin{tabular}{|l|l|l|}
\hline
\hline
\textbf{Feature Extractor Method}     &  \textbf{Acronym} \\
\hline
\hline
Color Layout       &  CL      \\ \hline
Scalable Color       & SC      \\ \hline
Color Structure   & CS      \\ \hline
Color Temperature   & CT      \\ \hline
Edge Histogram     & EH      \\ \hline
Texture Browsing    & TB      \\ \hline
\end{tabular}
\end{center}
\label{tab:acr}
\end{table}

\begin{table}[!htb]
\caption{Evaluation Function acronyms used in the experiments.}
\begin{center}
\begin{tabular}{|l|l|l|}
\hline
\hline
\textbf{Evaluation Function Name}     &  \textbf{Acronym} \\
\hline
\hline
City-Block       &  CB      \\ \hline
Euclidean       & EU      \\ \hline
Chebyshev      & CH      \\ \hline
Canberra     & CA      \\ \hline
\begin{tabular}[c]{@{}l@{}}Kullback Leibler\\ Divergence\end{tabular}    & KU      \\ \hline
Jeffrey Divergence    & JF \\ \hline
\end{tabular}
\end{center}
\label{tab:acr2}
\end{table}

\begin{figure*}[!htb]
\centering
\includegraphics[scale=1]{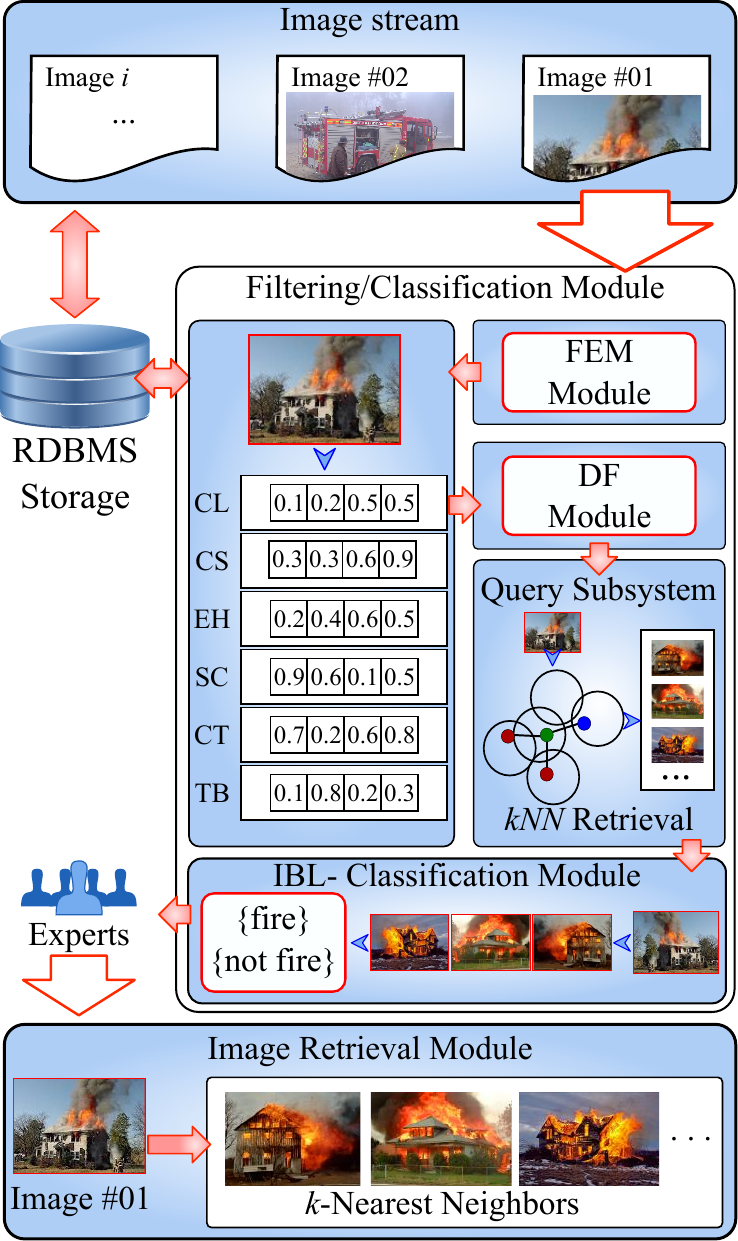}
\caption{Architecture of the $\FFDnR$. The Evaluating Module receives an unlabeled image, represents it executing feature extractor methods and labels it by using the Instance Based Learning Module. The system output (image plus label) interacts with the experts, who may also perform a similarity query.}
\label{fig:arch}
\end{figure*}

		Here we introduce the Fast-Fire Detection ($\FFDnR$) architecture, which uses image descriptors for image retrieval and classification.
		The architecture is organized in modules that implement the concepts reviewed in Section \ref{sec:background}.
		Figure \ref{fig:arch} illustrates the relationship among the modules, their communication and how they relate to a relational database management system (RDBMS). 

		The feature extraction methods module (the FEM module) accepts any kind of feature extractor. 
		For this work, we implemented six extractors following the MPEG-7 standard: Color Layout, Scalable Color, Color Structure, Edge Histogram, Color Temperature, and Texture Browsing -- explained in Section \ref{subsec:mpeg7fems}. 
		The evaluation functions module (the EF module) is also prepared for general implementations; for this work, we implemented six functions: City-Block, Euclidean, Chebyshev, Jeffrey Divergence, Kullback-Leibler Divergence, and Canberra.
		The  feature extractors methods and evaluation functions acronyms are listed in Tables \ref{tab:acr} and \ref{tab:acr2}.

		The architecture also has an Instance-Based Learning module (the IBL module), which classifies images labeling them as in classes $\{\mbox{\texttt{fire}}, ~\mbox{\texttt{not fire}}\}$.
		The IBL module receives as input the unclassified images, one image descriptor (a pair of feature extractor and evaluation function) and the set of past cases correctly labeled.
		This module is assisted by a similarity retrieval subsystem, which executes the $kNN$ queries necessary for the instance-based learning.

		We assume a flow of images is feed to Fast-Fire Detection architecture.
		As each image arrives, it is stored in the RDBMS along with the corresponding vectors of the features extracted.
		Then, the IBL module classifies each unlabeled image based on the required descriptor.

		This architecture was implemented as an API integrated to an RDBMS, in which the user can create his/her own image descriptor by combining FEM and DF to perform image classification.
		The user employs an SQL extension as the front-end for the architecture.
		The extension is able to execute similarity retrieval (through the Image Retrieval Module) and classification.

\section{\uppercase{Experiments}}
\label{sec:experiments}
\noindent 
		In this section, we search the combination of classifiers and image descriptors that are the most suitable to $\FFDnR$ in the fire detection task.
		We evaluate the impact of the image descriptors creating a candidate set of 36 descriptors given by the combination of the 6 feature extractors with the 6 evaluation functions considered in this work executing the IB1 classifier over the \texttt{\FlickrFire} dataset.
		The experiments were performed using the following procedure:
				\begin{enumerate}
		\item Calculate the F-measure metric to evaluate the efficacy of the experimental setting;
		\item Select the top-six image descriptors according to the F-measure to generate Precision-Recall plots, bringing more details about the behavior of the techniques;
		\item Validate our partial findings using Principal Component Analysis to plot the feature vectors of the extractors;
		\item Employed the top-three image descriptors according to the previous measures to perform a ROC curve evaluation, providing the analysis about the most accurate $\FFDnR$ setting;
		\item Evaluate the efficiency of the proposed $\FFDnR$ architecture, measuring the wall-clock time considering the multiple configurations of the descriptors.
		\end{enumerate}

\subsection{Obtaining the F-measure}\label{sec:f_measure}
\label{subsec:imagedesc}
\noindent
		To determine the most suitable $\FFDnR$ setting, we employed the F-measure, which relies on measuring the number of true positives (TP), false positives (FP) and false negatives (FN).
		The TP are the images containing fire which are correctly labeled as 'fire', while the FN are those labeled as 'not fire' although being fire images.
		The FN are the images labeled as 'fire' but containing no traces of fire.
		The F-measure is given by $2*TP/(2*TP + FP + FN)$.		
		
		We calculated the F-measure for the 36 image descriptors using 10-fold cross validation. 
		That is, for each round of evaluation, we used one tenth of the dataset to train the IB1 classifier and the remaining data for tests.
		It is performed 10 times and them the average F-measure is calculated.

		Table \ref{tab:results} presents the F-measure values for all the 36 combinations of feature extractor/evaluation function.
		The highest values obtained for each row are highlighted in bold.
		The experiment revealed that distinct descriptor combinations impact on fire detection.
		More specifically, the accuracy of extractors based on color is better than that of the extractors based on texture (Edge Histogram, and Texture Browsing). 
		Moreover, the extractors Color Layout and Color Structure have shown the best efficacy for fire detection, in combination respectively with the evaluation functions Euclidean and Jeffrey Divergence.
		
		The highlighted values are pointed out as the best setting for tuning $\FFDnR$.
		In addition, notice that the best descriptor achieved an accuracy of 85\%, which is close to the human labeling process, whose accuracy was 92.8\%.

\begin{table}[ht] 
\caption{F-Measure for each pair of feature extractor method (rows) versus evaluation function (columns). 
For each feature extractor, the evaluation function with the highest F-Measure is highlighted.} 
\centering 
 \begin{tabular}{| p{0.7cm}  | p{0.6cm}  p{0.6cm}  p{0.6cm}  p{0.6cm}  p{0.6cm}  p{0.6cm} |} 
\hline
\hline
 & \multicolumn{6}{|c|}{\textbf{Evaluation Functions}} \\

   \textbf{FEM} & \textbf{CB} & \textbf{EU} & \textbf{CH} & \textbf{CA} & \textbf{KU} & \textbf{JF} \\
\hline
\hline

\textbf{CL} & 0.834 & \textbf{0.847} & 0.807 & 0.828 & 0.803 & 0.844 \\ 
\textbf{SC} & \textbf{0.843} & 0.827 & 0.811 & 0.835 & 0.671 & 0.798 \\ 
\textbf{CS} & 0.853 & 0.849 & 0.821 & 0.848 & 0.746 & \textbf{0.866} \\ 
\textbf{CT} & 0.799 & 0.798 & 0.798 & \textbf{0.800} & 0.734 & 0.799  \\\hline
\textbf{EH} & 0.808 & 0.806 & 0.795 & 0.806 & 0.462 &  \textbf{0.815} \\ 
\textbf{TB} & \textbf{0.766} & 0.762 & 0.745 & 0.751 & 0.571 & 0.755  \\ 
 \hline
\end{tabular}
\label{tab:results}
\end{table}

		We also compared the best combination achieved by the IB1 classifier, as reported in Table 4, with other classifiers. 
		This was performed to assure that the instance-based learning (the $\FFDnR$ approach) is the most adequate classification strategy .
		In these experiments, we tuned $\FFDnR$ to employ the best EF for each FEM, as reported in Table \ref{tab:results}.
		We also grouped the results according to the employed FEM.
		
		Table \ref{tab:results_others} shows the $\FFDnR$ results compared to Naive-Bayes, J48, and RandomForest classifiers.
		The results show that $\FFDnR$ achieved the best F-Measure in every but one, of the classification configurations.
		Random Forest classification using the Scalable Color extractor beat $\FFDnR$, although by a narrow F-Measure margin.
		Thus, we can conclude that IB1 is adequate to fulfill the classification purpose on $\FFDnR$.

\begin{table}[ht] 
\caption{$\FFDnR$ obtained the highest F-Measure for all but one FEM when compared to other classifiers. For each feature extractor, we highlighted strategy with the highest F-Measure.} 
\centering 
 \begin{tabular}{| p{0.7cm}  | p{1.1cm} | p{1.1cm}  p{0.9cm}  p{1.3cm}  |} 
\hline
\hline
& \multicolumn{4}{c|}{\textbf{Classifiers}} \\ \cline{2-5}
\textbf{FEM} & \textbf{FFireDt} & \textbf{Naive-Bayes} & \textbf{J48} & \textbf{Random Forest} \\
\hline
\hline

\textbf{CL} & \textbf{0.847} & 0.787 & 0.751 & 0.829 \\ 
\textbf{SC} & 0.843 & 0.808 & 0.845 & \textbf{0.864} \\ 
\textbf{CS} & \textbf{0.866} & 0.406 & 0.842 & 0.866 \\ 
\textbf{CT} & \textbf{0.800} & 0.341 & 0.800 & 0.774 \\ 
\textbf{EH} & \textbf{0.815} & 0.522 & 0.711 & 0.787 \\ 
\textbf{TB} & \textbf{0.766} & 0.476 & 0.706 & 0.723 \\ 
 \hline
\end{tabular}
\label{tab:results_others}
\end{table}

\begin{figure*}[hbt!]
\centering
\includegraphics[scale=0.6]{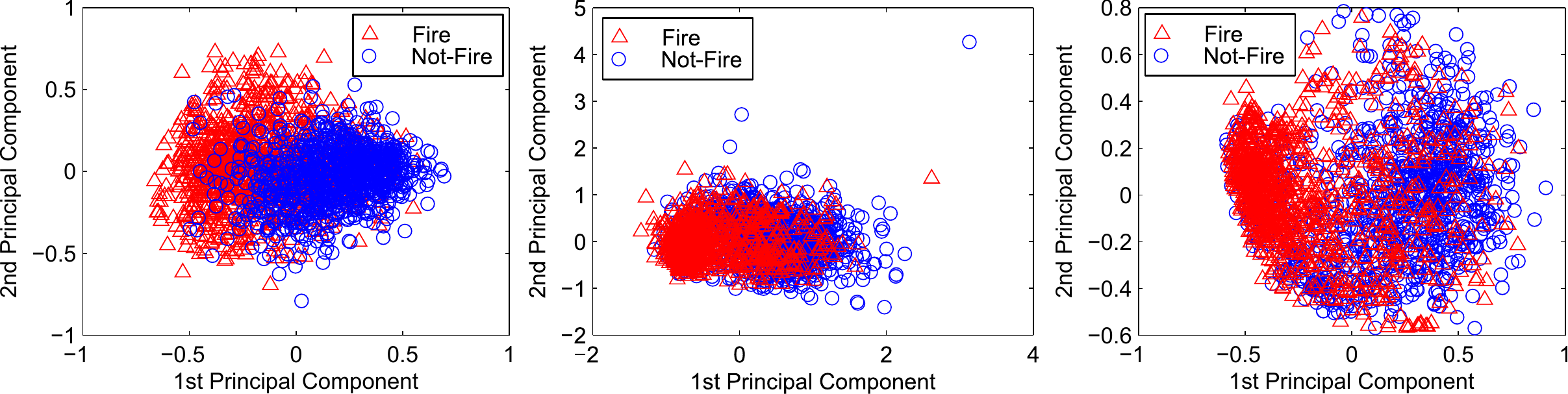}
\caption{PCA projection of fire and not-fire images: (a) Color Layout, (b) Color Structure, (c) Scalable Color and (d) Edge Histogram. The Color Layout visually separates the dataset into two clusters.}
\label{fig:plot_a}
\end{figure*}

\subsection{Precision-Recall}
\label{subsec:PR}
\noindent

\begin{figure}[b]
\centering
\includegraphics[scale=0.86]{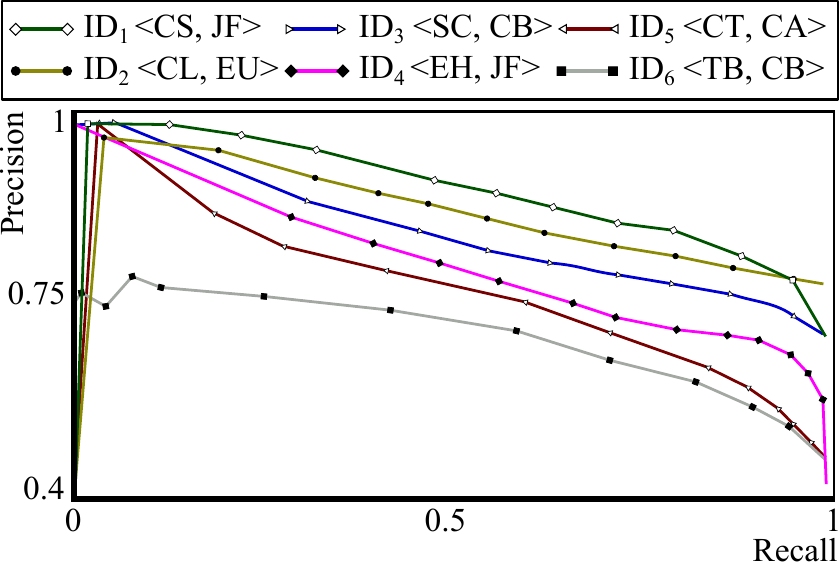}
\caption{{\it Precision vs. Recall} graphs for each of the most precise image descriptor combinations according to F-measure.}
\label{fig:pr}
\end{figure}

		We extended the analysis of Section \ref{subsec:imagedesc} measuring the Precision and Recall of each configuration, which are values also employed on F-measure.
		Such further analysis permits to better understand the behavior of the image descriptors and more specifically, the behavior of the feature extractors.	
		A Precision vs. Recall (P$\times$R) curve is suitable to measure the number of relevant images regarding the number of retrieved elements.
	We used Precision vs. Recall as a complementary measure to determine the potential of each image descriptor in the $\FFDnR$ setting.
	A rule of thumb on reading P$\times$R curves is: the closer to the top the better the result is.
	Accordingly, we consider only the more efficient combination of each feature extractor, as highlighted in Table \ref{tab:results}: 
				$ID_1<$CS, JF$>$, 
			  $ID_2<$CL, EU$>$, 
				$ID_3<$SC, CB$>$,
				$ID_4<$EH, JF$>$,
				$ID_5<$CT, CA$>$ and 
				$ID_6<$TB, CB$>$.
		Figure \ref{fig:pr} confirms that the image descriptors $ID_1$, $ID_2$, and $ID_3$ are in fact the most effective combinations for fire detection.
		It also shows that, for those three descriptors, the precision is at least 0.8 for a recall of up to 0.5, dropping almost linearly with a small slope, which can be considered acceptable. 
		This observation reinforces the findings of the F-measure metric, indicating that the behavior of the descriptors are homogeneous and well-suited for the task of retrieval and, consequently, for classification purposes.

\subsection{Visualization of feature extractors}
\label{sec:visualization}
		Based on the results shown so far, we hypothesize that the Color Structure, Color Layout, and Scalable Color extractors are the most adequate to act as $\FFDnR$ setting.
		In this section, we look for further evidence, using visualization techniques to better understand the feature space of the extractors using Principal Component Analysis (PCA).
		The PCA analysis takes as input the extracted features, which may have several dimensions according to the FEM domain, and reduces them to two features.
		Such reduction allows us to visualize the data as a scatter-plot.
		
		Our hypothesis shall gain more credibility if the corresponding visualizations allow seeing a best separation of the classes {\it fire} and {\it not-fire}, in comparison to the other three extractors.
		Figures \ref{fig:plot_a}(a) - \ref{fig:plot_a}(d) allow visualizing the two-dimensional projection of the data, plotting the two principal components of the PCA processing of the space generated by each extractor.
		Figure \ref{fig:plot_a}(a) depicts the representation of the data space generated by CL, the extractor that presented the better separability in the classification process.
		The two clusters can be seen as two well-formed clouds with a reasonably small overlapping, splitting the images as containing fire or not.
		
		Figure \ref{fig:plot_a}(b) shows the data visualization of the space generated by CS, which was the FEM that obtained the highest F-measure on previous experiments.
		The data projection shows that each cluster forms a cloud clearly identifiable, having the centers of the clouds distinctly separated.
		However, this figure reveals that there is a large overlap between the two classes.
		Figure \ref{fig:plot_a}(c) presents the projection of the space generated by the SC extractor. 
		Again, it can be seen that there are two clusters, but with an even larger overlap between them.
		Visually, the CL outperformed the other color FEMS: the CS and SC, when drawing the border between the two classes.
		
		Figure \ref{fig:plot_a}(d) depicts the visualization generated by the EH extractor.
		It can be seen that indeed it has two clouds: one almost vertical to the left and another along the ``diagonal'' of the figure.
		However, the two clouds are not related to the existence of fire, as the elements of both clusters are distributed over both clouds.
		Figures \ref{fig:plot_a} (a), \ref{fig:plot_a} (b) and \ref{fig:plot_a} (c) show the visualization of the extractors based on color.
		The four visualizations show that the corresponding CL, SC and SC indeed generate clusters.
		However, there are increasing larger overlaps between fire and not-fire instances.
		Regarding the TB and CT features, the PCA projection was not able to separate the fire and not-fire classes.
				Concluding, the visualization of the feature spaces shows that extractors based on color are able to separate the data into visual clouds related to the expected clusters. 
		Particularly, Color Layout has shown the best visualization, followed by Color Structure, and Scalable Color, which also have shown to significantly separate the classes.
		However, the extractors based on texture identify characteristics that are not related to fire, thus presenting the worst separability, as expected.

\subsection{ROC curves}\label{subsec:roc}
\noindent

\begin{figure}[ht]
\centering
\includegraphics[scale=0.86]{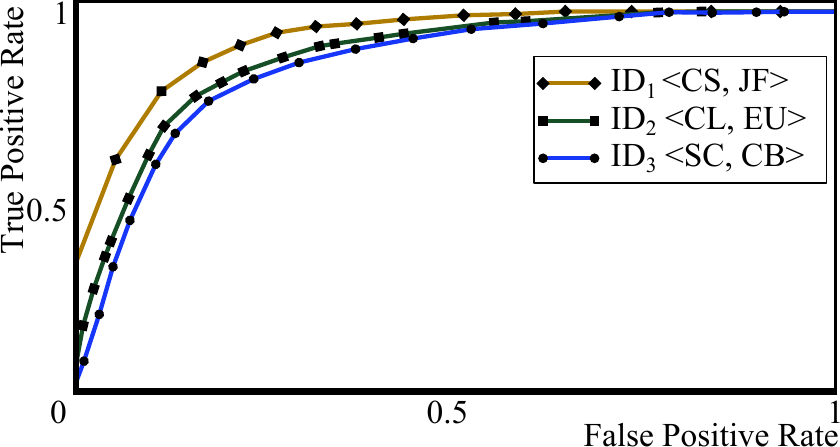}
\caption{ROC curves for the top three image descriptors in the task of fire detection.}
\label{fig:roc_a}
\end{figure}

		Finally, we employed one last accuracy measure to define the $\FFDnR$ setting: the ROC curve.
		It allow us to determine the experiments overall accuracy, using measures of sensitivity and specificity.
		Figure \ref{fig:roc_a} presents the detailed ROC curves for image descriptors $ID_1 = <$CS, JF$>$, $ID_2 = <$CL, EU$>$, and $ID_3 = <$SC, CB$>$, the top three best combinations according to the F-Measure, Precision-Recall and Visualization experiments.
		For fire-detection, the area under the ROC curve was up to 0.93 for $ID_1$; up to 0.87 for $ID_2$; and up to 0.85 for $ID_3$.

		These results indicate that the top three image descriptors have similar and satisfactory accuracy.
		Therefore, the choice of which descriptor to use becomes a matter of performance.
		In the next section we evaluate the performance to conclude what is the most adequate descriptor.

\subsection{Processing Time and Scalability}\label{sec:scalability}
\noindent{When monitoring images originated from social media, the time constraint is important because of the high rate at which new images arrive. 
Thus, we also evaluate the efficiency, given in wall-clock time, of the candidate image descriptors. We ran the experiments in a personal computer equipped with processor Intel Core i7 R 2.67 GHz with 4GB memory over operating system Ubuntu 14.04 LTS.}\\

\noindent{\bf Feature Extractors}\\
\noindent

\begin{figure}[ht]
\centering
\includegraphics[scale=0.86]{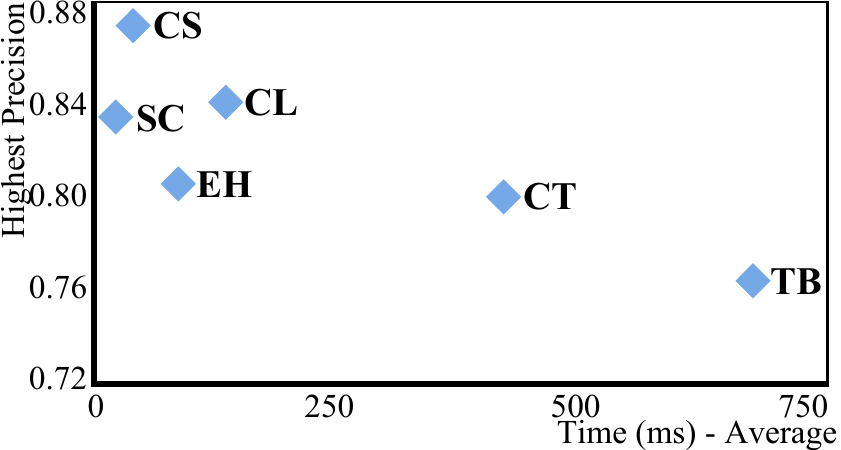}
\caption{Plot {\it Highest precision when classifying dataset \texttt{\FlickrFire} vs. Average time to extract the features of one image} for the six feature extractors.}
\label{fig:time_fem}
\end{figure}

		When monitoring images originated from social media, the time constraint is important because of the high rate at which new images may arrive. 
		Thus, we also evaluate the efficiency, given in wall-clock time, of the candidate image descriptors.
		Figure \ref{fig:time_fem} shows the required average time to perform the feature extraction on $\FFDnR$ regarding \texttt{\FlickrFire} dataset.
  	Color Structure, the most precise extractor, was the second fastest.
		The second and third most precise extractors were Color Layout and Scalable Color: the former was three times slower than Color Structure, and the later was the fastest extractor.
		Thus, we are now able to state the that extractors Color Structure and Scalable Color are the best choices for fire detection in image streams.
		Meanwhile, the texture-based extractors Edge Histogram, and Texture Browsing presented low performances, so they are definitely dismissed as possible choices.\\

\begin{figure}[ht]
\centering
\includegraphics[scale=0.86]{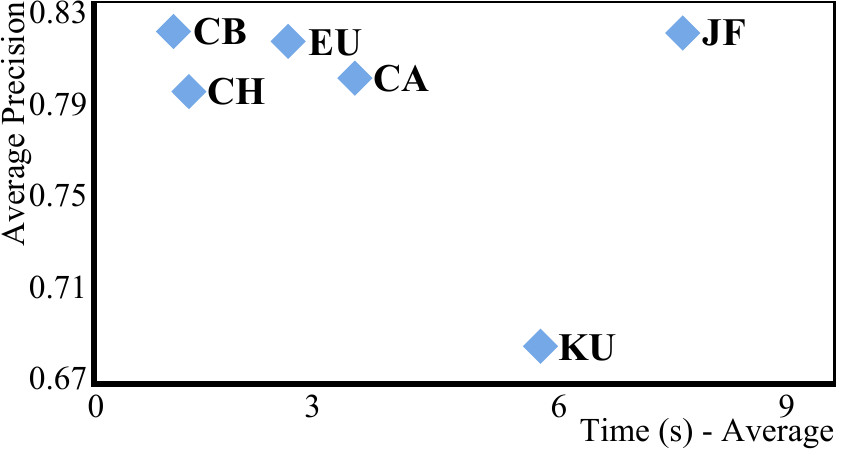}
\caption{Plot {\it Average precision when classifying dataset \texttt{\FlickrFire} vs. Time to perform 2 trillion calculations} for the six evaluation functions.}
\label{fig:time_df}
\end{figure}

\noindent{\bf Evaluation Functions}\\
\noindent
		Figure \ref{fig:time_df} shows the time required to perform 2 trillion evaluation calculations for each evaluation function on feature vectors of 256 dimensions.
		The plot {\it average precision vs. wall-clock time} shows that, although the Jeffrey Divergence demonstrated the highest precision, it was the least efficient.
		In their turn, the City-Block and Euclidean distances presented excellent performance and a precision only slightly below the Jeffrey Divergence. 
		Therefore, we can say that they are the most adequate evaluation functions when performance is on concern, such as is the case in our problem domain.
		Finally, we conclude that the image descriptors given by the combinations $\{CS,SC\} \times \{CB,EU\}$ are the best options in terms of both efficacy (precision) and efficiency (wall-clock time). 
		In Table \ref{tab:resultsfinal} we reproduce the F-measure results highlighting the most adequate combinations according to our findings.

\begin{table}[ht] 
\caption{F-Measure for each pair of feature extractor method (rows) and evaluation function (columns), now highlighting the best combinations according to our experiments.} 
\centering 
 \begin{tabular}{| p{0.7cm}  | p{0.6cm}  p{0.6cm}  p{0.6cm}  p{0.6cm}  p{0.6cm}  p{0.6cm} |} 
\hline
\hline
 & \multicolumn{6}{|c|}{\textbf{Evaluation Functions}} \\

   \textbf{FEM} & \textbf{CB} & \textbf{EU} & CH & JF & KU & CA \\
\hline
\hline

CL & 0.834 & 0.847 & 0.807 & 0.844 & 0.803 & 0.828 \\
\textbf{CS} & \textbf{0.853} & \textbf{0.849} & 0.821 & 0.866 & 0.746 & 0.848 \\ 
\textbf{SC} & \textbf{0.843} & \textbf{0.827} & 0.811 & 0.798 & 0.671 & 0.835 \\
EH & 0.808 & 0.806 & 0.795 & 0.815 & 0.462 & 0.806 \\ 
CT & 0.799 & 0.798 & 0.798 & 0.799 & 0.734 & 0.800 \\ 
TB & 0.766 & 0.762 & 0.745 & 0.755 & 0.571 & 0.751 \\ 
 \hline
\end{tabular}
\label{tab:resultsfinal}
\end{table}

By these experiments, we point out that the best image descriptor for fire detection, considering the built dataset, is given by the combinations of the MPEG-7 Color Structure and Scalable Color extractors with the distance functions City-Block and Euclidean.
		These combinations provide not only more efficacy, but also more efficiency.
		In general, we noticed that feature extractors based on color were more effective than extractors based on texture.
		We also identified that the Jeffrey divergence was the most accurate, however, it was also the most expensive evaluation function.

\section{\uppercase{Conclusions}}\label{sec:conclusion}
\noindent
		We worked on the problem of identifying fire in social-media image sets in order to assist rescue services during emergency situations.
		The approach was based on an architecture for content-based image retrieval and classification.
		Using this architecture, we compared the accuracy and performance (processing time) of image descriptors (pairs of feature extractor and evaluation function) in the task of identifying fire in the images.
		As a ground-truth, we built a dataset with 2,000 human-annotated images obtained from website Flickr.
		Our contributions in this paper are summarized as follows:

\begin{enumerate}
\item \textbf{Dataset \FlickrFire:} we built a varied human-annotated dataset of real images suitable
as ground-truth to foster the development of more precise techniques for automatic identification of fire;

\item \textbf{Fast-Fire Detection (\FFDnR) Architecture:} we designed and implemented a flexible, scalable, and accurate method for content-based image retrieval and classification to be used as a model for future developments in the field;

\item \textbf{Evaluation of existing techniques:} we compared 36 combinations of MEPG-7 feature extractors with evaluation functions from the literature considering their potential for accurate retrieval using F-measure and Precision-Recall. As a result, we achieved 85\% accuracy, precise classification (ROC curves), and efficient performance (wall-clock time).
\end{enumerate}

		Our results showed that $\FFDnR$ was able to achieve a precision for the fire detection task, which is comparable to that of human annotators.
		We conclude by stressing the importance of monitoring images from social media, including situations in which decision making can benefit from precise and accurate information.
		However, to take advantage of them, it is required to have automated tools that can flag them from the social media as soon as possible, and this work is an important step toward it.

\section*{ACKNOWLEDGEMENTS}
This research is supported, in part, by FAPESP, CNPq, CAPES, STIC-AmSud, the RESCUER project, funded by the European Commission (Grant: 614154) and by the CNPq/MCTI (Grant: 490084/2013-3).

\bibliographystyle{apalike}

\vfill
\end{document}